# The Complexity of Integer Bound Propagation

**Lucas Bordeaux**                                               LUCASB@MICROSOFT.COM
*Microsoft Research, 7 J J Thomson Avenue, CB30FB*
*Cambridge, UNITED KINGDOM*

**George Katsirelos**                                            GKATSI@GMAIL.COM
*LRI, Université Paris-Sud 11*
*Paris, FRANCE*

**Nina Narodytska**                                              NINAN@CSE.UNSW.EDU.AU
*NICTA Neville Roach Laboratory and*
*University of New South Wales*
*223 Anzac Parade Kensington NSW 2052, AUSTRALIA*

**Moshe Y. Vardi**                                               VARDI@CS.RICE.EDU
*Rice University, P. O. Box 1892*
*Houston, TX 77251-1892, U.S.A.*

## Abstract

Bound propagation is an important Artificial Intelligence technique used in Constraint Programming tools to deal with numerical constraints. It is typically embedded within a search procedure ("branch and prune") and used at every node of the search tree to narrow down the search space, so it is critical that it be fast. The procedure invokes constraint propagators until a common fixpoint is reached, but the known algorithms for this have a *pseudo*-polynomial worst-case time complexity: they are fast indeed when the variables have a small numerical range, but they have the well-known problem of being prohibitively slow when these ranges are large. An important question is therefore whether *strongly*-polynomial algorithms exist that compute the common bound consistent fixpoint of a set of constraints. This paper answers this question. In particular we show that this fixpoint computation *is in fact NP-complete*, even when restricted to binary linear constraints.

## 1. Introduction and Overview of the Main Results

Constraint solvers typically solve problems by interleaving search and propagation. Propagation is an iterative procedure which, at each iteration, *propagates* every constraint in the problem to narrow the domains of its variables. The iteration stops when no constraint changes the domains of its variables. In this case, propagation has reached a *common fixpoint* for all constraints. This iterative algorithm is guaranteed to compute the fixpoint in polynomial time if propagating each constraint takes polynomial time and the domains of the variables are defined as lists of values. Very often, however, it is inconvenient or infeasible to list all values explicitly: instead the domains are defined by lower and upper bounds. We focus on this representation, and on variables taking *integer* values. In this setting, computing a fixpoint by the iterative algorithm may require exponential time even if each constraint can be propagated in polynomial time. We show that this exponential behaviour is not simply due to the iterative algorithm being suboptimal; rather it is intrinsic to the problem of computing a fixpoint, as it is NP-complete even when the system of constraints is restricted to binary linear inequality constraints.

## 1.1 Bound Propagation and Slow Convergence

We illustrate the behaviour of the iterative fixpoint algorithm using a system of two constraints:

$$x + y = 7, \quad x + 1 \geq 2y, \qquad \text{with initial bounds:} \qquad x \in [0,5], \quad y \in [0,10]$$

A possible trace of the fixpoint computation is the following. The lower bound of $y$ is initially 0 but from the constraint $x + y = 7$ we deduce that $y$ cannot take values 0 or 1: if it does, then the sum is $< 7$, *even if we fix $x$ to its highest allowed value.* Therefore the intervals can be narrowed down to $x \in [0,5], y \in [\underline{2}, 10]$. Similarly:

| | | | |
|---|---|---|---|
| from | $x + y = 7,$ | we deduce: | $x \in [0,5], y \in [2, \underline{7}]$; |
| from | $x + 1 \geq 2y,$ | we deduce: | $x \in [\underline{3}, 5], y \in [2, 7]$; |
| | | and: | $x \in [3,5], y \in [2, \underline{3}]$; |
| back to | $x + y = 7,$ | we now deduce: | $x \in [\underline{4}, 5], y \in [2, 3]$. |

At this point we have reached a common fixpoint for both constraints, because we cannot deduce that the domains need to be narrowed any further.

This algorithm, however, exhibits *slow convergence* behaviour even in deceivingly simple examples such as:

$$x < y, \quad y < x \qquad \text{with initial bounds:} \qquad x \in [0, 10^8], \quad y \in [0, 10^8] \tag{1}$$

The iterative algorithm for fixpoint computation shrinks the bounds by one unit in each iteration, which means that $10^8$ iterations will be required to reach the fixpoint, which in this case is empty. This slow convergence is in fact *exponential* in the size of the problem representation, as $\log(10^8)$ bits are enough to represent each bound. This behaviour is not limited to artificial examples as the previous one but in fact happens time and again when solving problems with large numerical ranges. This severely limits the application of CP in areas such as software verification or theorem proving where large ranges are needed (e.g., the whole 32-bit integer range).

Due to the importance of the problem, efforts have been made to alleviate slow convergence, notably Jaffar, Maher, Stuckey, and Yap (1994), Lhomme, Gottlieb, Rueher, and Taillibert (1996), Lebbah and Lhomme (2002), Leconte and Berstel (2006); but all proposed algorithmic improvements prevent slow convergence only in specific cases. Fully addressing the slow convergence problem would require a strongly polynomial algorithm for fixpoint computation. Therefore the question is: does such an algorithm exist, or is bound propagation in fact intractable?

## 1.2 Prior Complexity Results on Propagation

Standard propagation algorithms are iterative processes that apply "propagators", i.e., narrowing functions associated with each constraint, until reaching a fixpoint. Their complexity is therefore determined by two complementary questions:

**Q1:** How hard is it to compute each propagator?

**Q2:** How hard is it to find a common fixpoint of the propagators?

The complexity of constraint propagation has in a sense been extremely well-studied, but all the results we are aware of for bound propagation deal with Question 1 only. Those prior hardness results showed that for some complex constraints we cannot have polynomial-time propagators reaching certain levels of consistency. Two such results are:

- Given a linear equality it has been observed (Yuanlin & Yap, 2000; Choi, Harvey, Lee, & Stuckey, 2006) that any propagator that reaches arc consistency or bound(Z) consistency[1] needs to solve a knapsack problem, which is NP-complete in the weak sense. For this reason propagators for linear constraints used in practice either reach a weaker consistency such as bound(R) consistency, or are restricted to very small domains, as proposed for instance in Trick (2001).

- Results by Bessière (2006) prove that even bounded-arity (two-variable) constraints can be constructed for which checking bound(Z) consistency is NP-complete.

Question 2 only makes sense, of course, in the common case where the propagators are polynomial-time computable (if they are not, computing their common fixpoint cannot be easy in general). The only known fact in this case is that the standard, iterative propagation algorithms often take an exponential number of steps to reach a fixpoint in practice, as mentioned and illustrated in Section 1.1. This leaves open the question whether better algorithms exist or fixpoint computation is, in fact, intrinsically hard.

## 1.3 Our Main Results

In this paper we consider very simple, common propagators and address Question 2. We show that in general *even surprisingly simple propagators can lead to a fixpoint computation problem that is NP-hard.* This not only explains why all standard, iterative fixpoint-computation algorithms have an exponential worst-case in practice, but also shows it is unlikely that the there exists an algorithm with better worst case. In particular an important class of simple propagators whose fixpoint computation is NP-hard is the bound(R) consistency propagators for linear constraints (Proposition 1). These are ubiquitous constraints, and very weak and widely used propagators for these constraints. Many problems that use numerical computations and large domains tend to include at least linear constraints, therefore there are few cases where slow convergence will be avoidable. We nevertheless identify one such case: if the coefficients of the linear constraints are all unit (1 in absolute value), then bound(R) consistency can be obtained in polynomial time by a non-standard propagation algorithm based on Linear Programming. We also study other types of basic numerical constraints: multiplication and max.

## 1.4 Outline

In **Section 2** we summarize the required material on Constraint Satisfaction Problems and bound propagation. **Section 3** then focuses on linear constraints. We prove the aforementioned Proposition 1, then identify restricted forms of linear constraints for which propagation is tractable. **Section 4** presents results for some other basic propagators: for quadratic constraints, the hardness result can be strengthened and holds even for a fixed number of variables; for max constraints, fixpoint computation has an interesting complexity (between P and NP-complete) and is proved equivalent to an important open problem; we last comment on max-closed constraints. We conclude in **Section 5**.

---

1. We give the formal definitions of bound(R) and bound(Z) consistency in Section 2.2.

## 2. Formal Background

In this section we summarize the required material on Constraint Satisfaction Problems and bound propagation. More details on this material can be found in papers by e.g. Schulte and Carlsson (2006), and Bessière (2006).

### 2.1 Constraint Satisfaction Problems

A Constraint Satisfaction Problem (CSP) is a triple $\langle X, D, C \rangle$, where: $X = \{x_1 \cdots x_n\}$ is a set of variables, $D = \{D_1 \cdots D_n\}$ is a set of finite domains (finite sets of values), one for each variable, and $C$ is a set of constraints. In this paper we consider discrete domains: all the elements in $D$ are *integers*. For the moment we simply define constraints very generally as logical predicates over subsets of $X$; later in the paper we consider specific types of constraints, for instance linear ones. An *assignment* is a function $\tau$ that assigns a value $\tau(x_i) \in D_i$ to every variable $x_i$. A *solution* to the CSP is an assignment that satisfies the constraints. Throughout the paper, we keep the following conventions:

- $n = |X|$ denotes the number of variables;

- $m = |C|$ is the number of constraints;

- $d = \max_{i \in 1..n} |D_i|$ is the size of the largest domain.

It is important to note that $D_i$ may be represented as an interval, rather than an explicit set of values. In this work, we only consider domains represented as intervals: each domain is of the form $D_i = [l_i, u_i]$, where $l_i$ and $u_i$ are the lower and upper bounds of the domain.

### 2.2 Propagators and Notions of Bound Consistency

The constraints of the problem are associated with *propagators*. (In our setting there will be, in general, several propagators per constraint.) We follow the classical presentation of propagators as operators on a lattice, initiated in work by Benhamou (1996) and on which more details can be found in papers by Apt (1999), and Schulte and Carlsson (2006): each propagator is a function that can narrow the domains of (some of) the variables, removing values that cannot appear in any solution. Thus, we talk about the *current* domain of a variable $x_i$, as the result of it being narrowed by the application of one or more propagators. We denote by $x_i^-$ the current lower bound of $x_i$ and by $x_i^+$ its current upper bound. $x_i^-$ and $x_i^+$ are initially set to the initial bounds $l_i, u_i$ and remain afterwards constrained by $l_i \leq x_i^- \leq x_i^+ \leq u_i$. We denote by $\mathcal{D}$ the Cartesian product of the intervals $[l_i, u_i]$ for all $i \in 1..n$.

**Definition 1 (Propagator)** *A propagator for a constraint $k \in 1..m$ is a function $f : \mathcal{P}(\mathcal{D}) \to \mathcal{P}(\mathcal{D})$, that is:*

- *monotone, i.e., $A' \subseteq A \to f(A') \subseteq f(A)$;*

- *contracting, i.e., $f(A) \subseteq A$;*

- *correct, i.e., no point in $A \setminus f(A)$ satisfies the constraint.*

*We restrict ourselves to propagators that are* polynomial-time computable. *Bound consistency propagators are additionally restricted to elements of $\mathcal{P}(\mathcal{D})$ that are representable as Cartesian products of intervals, plus the special value $\emptyset$.*

Several types of propagators can be used for numerical constraints; these propagators are characterized by the *level of consistency* they enforce. Since we have restricted our focus to interval domains, we present only *bound consistency*. The two main variants are bound(Z) and bound(R) consistency:

**Definition 2 (Bound(Z|R) support)** *A bound(Z) (bound(R)) support of a constraint $k$ is an assignment $\tau$ of integer (real) values to the variables $X$ such that $x_i^- \leq \tau(x_i) \leq x_i^+$ for $i \in 1..n$ and $\tau$ satisfies the constraint $k$.*

**Definition 3 (Bound(Z|R) consistency)** *A constraint $k$ is bound(Z) (bound(R)) consistent iff for every variable $x_i \in X$, there exists a bound(Z) (bound(R)) support $\tau^-$ with $\tau^-(x_i) = x_i^-$ and a bound(Z) (bound(R)) support $\tau^+$ with $\tau^+(x_i) = x_i^+$.*

The difference between the two is easily understood on an example:

**Example 1** *Consider the constraint $2x + 2y + 3z = 4$.*

- *The intervals $x, y, z \in [0, 1]$ are bound(R) consistent since each of the integer bounds has a real-valued* support: $x = 0$ is supported by the tuple $(x = 0, y = 1, z = 2/3)$; $z = 1$ and $y = 0$ by the tuple $(x = 1/2, y = 0, z = 1)$; $x = 1$, $y = 1$ and $z = 0$ by the tuple $(x = 1, y = 1, z = 0)$.*

- *These intervals are, however, not bound(Z) consistent: the only integer solution is $(x = 1, y = 1, z = 0)$, which means that bound(Z) consistency would reduce the bounds further to $x \in [1, 1], y \in [1, 1], z \in [0, 0]$.*

Bound(Z) consistency requires that we check the existence of an integer-valued support, and for some classes of constraints such as linear equalities each propagator would need to solve an NP-complete problem. Since our focus is on the computation of common fixpoint of *simple* operators, we only consider bound(R) consistency in this paper. As noted previously in the literature (Schulte & Stuckey, 2005), bound(R) consistency it is in fact "the bound consistency implemented for most primitive constraints", precisely because it is often the only one for which propagators are easy to compute in general for large domains. In the rest of the paper we focus on several of the main basic types of numerical constraints (in particular linear ones), and give further details on the bound(R) consistency propagators obtained for these constraints. In all the cases we consider the propagators are very simple indeed.

$A := \mathcal{D}$
$change := \mathbf{true}$
$\mathbf{while} \ change \ \mathbf{do}$
    $change := \mathbf{false}$
    $\mathbf{foreach} \ f \in F \ \mathbf{do}$
        $oldA := A$
        $A = f(A)$
        $\mathbf{if} \ A \neq oldA \ \mathbf{then} \ change := \mathbf{true}$
    $\mathbf{done}$
$\mathbf{done}$

Figure 1: A simple fixpoint computation algorithm.

## 2.3 Fixpoints

Propagators are monotone narrowing operators, thus we may consider the problem of identifying the greatest common fixpoint of a set of propagators.

**Definition 4 (Greatest Common Fixpoint)** *The greatest common fixpoint $\mathsf{gfp}(F)$ of a set of propagators $F$ is the largest Cartesian product of intervals $A \subseteq \mathcal{D}$ such that for each operator $f \in F$, we have $f(A) = A$.*

There are two computational problems related to fixpoints:

- *Function Problem:* Effectively compute $\mathsf{gfp}(F)$;

- *Decision Problem:* Decide whether $\mathsf{gfp}(F) \neq \emptyset$, i.e., whether there exists a (non-empty) fixpoint. (Note that our definition of propagators implies that $f(\emptyset) = \emptyset$ for all $f \in F$. Therefore $\emptyset$ is always a common fixpoint.) In other words: do the propagators stabilize to non-empty domains?

As often in complexity work we mostly focus on the Decision problem in this paper. The reason is that the basic complexity classes (NP in particular) are defined for decision problems, and that hardness results on the decision problem also imply that the function problem is hard. The only place where we refer to the function problem is this section, where we describe the basic greatest fixpoint computation algorithm.

An algorithm for computing $\mathsf{gfp}(F)$ is specified in Fig. 1. It is presented in its simplest form, which excludes several possible optimizations related, in particular, to the fact that not all constraints necessarily deal with all variables. (These optimizations are well-known and orthogonal to our discussion in this paper.) In this algorithm we initialize the Cartesian product of domains to $\mathcal{D}$, in other words we initially have $x_i^- = l_i$ and $x_i^+ = u_i$, for all $i \in 1..n$; and we simply apply all propagators until a stable state is reached, i.e., no propagator shrinks any domain further. The reader can verify that this algorithm specifies formally the reasoning that we presented informally in our introductory example (Sec. 1.1).

### 2.4 Complexity Upper Bound of Fixpoint Propagation

The worst-case time upper bound of fixpoint computation can be analyzed as follows[2]. Let $p = |F|$ be the number of propagators. (Note that we have in general one or more propagators per constraint, i.e., $p \geq m$.) We enter the *while* loop at most $nd$ times since at every new iteration we must reduce at least one bound by one unit, and each time the *foreach* loop is entered at most $p$ times. Overall the algorithm therefore terminates after a number of propagator applications of:

$$\mathcal{O}(npd).$$

In other words, it is in fact exponential in the number of bits of the encoding: this complexity can be written $\mathcal{O}(np \cdot 2^b)$, where $b$ is the number of bits of the bound encoding. This is despite the fact that each propagator is polynomial in the size of the encoding. Such algorithms are called *pseudo-polynomial*. In contrast algorithms that are truly polynomial in the number of bits of the encoding, i.e., whose worst-case time complexity is $\mathcal{O}(\pi(n, m, \log d))$, for some polynomial $\pi$, are called *strongly polynomial* (Papadimitiou, 1994). The problem of a pseudo-polynomial algorithm for this problem is that it scales linearly with the size of the domains, which may themselves be exponentially large. Since the propagators we consider take strongly polynomial time, the analyis of the upper bound is summarized as follows:

**Observation 1** *The naive fixpoint computation algorithm (Fig. 1) always terminates in* pseudo-*polynomial-time.*

The question is whether *strongly* polynomial algorithms exist. The rest of the paper focusses on this question, for several classes of propagators.

## 3. Linear Constraints

In this section we consider linear inequalities, i.e., our set of constraints $C$ contains $m$ inequalities of the form:

$$\sum_{i \in 1...n} a_{i,k} x_i \geq c_k, \quad k \in 1 \ldots m \tag{2}$$

where each $c_k$ and $a_{i,k}$ are integers. It is convenient to introduce some extra notation: we denote by $s_{i,k}$ the sign of the $i$th term in constraint $k$, i.e.,:

$$s_{i,k} = \begin{cases} + & if \quad a_{i,k} \geq 0 \\ - & if \quad a_{i,k} < 0 \end{cases} \tag{3}$$

Moreover, given a sign $s \in \{-, +\}$, the sign $-s$ is defined as $+$ if $s = -$ and as $-$ otherwise. The sign $+s$ will simply denote $s$. With this notation the terms $a_{i,k} x_i^{-s_{i,k}}$ and $a_{i,k} x_i^{+s_{i,k}}$ simply represent the smallest and largest elements of the set $\{a_{i,k} v \mid v \in [x_i^-, x_i^+]\}$.

---

2. Few papers give explicit upper bounds on the complexity of computing a fixpoint of a set of bound consistency propagators. The earliest reference we are aware of is the work of Lhomme (1993); it considers constraints on the reals but assumes finite precision (floating points), and its analysis directly adapts to discrete intervals.

### 3.1 Bound(R) Consistency Propagators for Linear Inequalities

We briefly summarize the material we need on bound(R) consistency in the case of linear inequalities. We refer the reader to the literature for more details, in particular the papers by Harvey and Stuckey (2003), Schulte and Carlsson (2006), Bessière (2006), and Apt and Zoteweij (2007) have substantial material on bound(R) consistency and linear constraints. Also of interest are works that show how to improve bound propagation for long linear constraints (Harvey & Schimpf, 2002; Katriel, Sellmann, Upfal, & Van Hentenryck, 2007).

Consider a variable $x_i$. Its bound $x_i^{-s_{i,k}}$ is *bound(R) inconsistent* w.r.t. the $k$th inequality of the system iff: *even when we fix the other terms to their maximum*, we obtain something lower than $c_k$. It is *bound consistent* if the opposite is true i.e., iff:

$$a_{1,k}x_1^{+s_{1,k}} + \ldots + a_{i-1,k}x_{i-1}^{+s_{i-1,k}} + \boxed{a_{i,k}x_i^{-s_{i,k}}} + a_{i+1,k}x_{i+1}^{+s_{i+1,k}} + \ldots + a_{n,k}x_n^{+s_{n,k}} \geq c_k \quad (4)$$

We call this the *bound consistency inequality* of variable $x_i$ w.r.t. constraint $k$. The bound consistency propagator for a linear inequality simply shrinks the bounds of each variable $x_i$. Let:

$$q_{i,k} = c_k - \sum_{j \in [1,n], \ j \neq i} a_{j,k}x_j^{+s_{j,k}}$$

be the minimal quantity that has to be reached by $a_{i,k}x_i^{-s_{i,k}}$ to satisfy the bound consistency inequality (in other words: $x_i$ is bound consistent w.r.t. constraint $k$ iff $a_{i,k}x_i^{-s_{i,k}} \geq q_{i,k}$). The (bound(R) consistency) propagator associated with constraint $k \in 1..m$ and variable $i \in 1..n$ is the function that reduces the bound of $x_i$ to the closest bound consistent value. It is defined by the following pseudo-code:

$$\boxed{L_{i,k}:} \quad \begin{aligned} &\text{if } a_{i,k} > 0 \text{ then} \quad x_i^- := \max\left(x_i^-, \left\lceil \frac{q_{i,k}}{a_{i,k}} \right\rceil\right) \\ &\text{if } a_{i,k} < 0 \text{ then} \quad x_i^+ := \min\left(x_i^+, \left\lfloor \frac{q_{i,k}}{a_{i,k}} \right\rfloor\right) \end{aligned} \quad (5)$$

(The propagator does nothing if $a_{i,k} = 0$.)

### 3.2 NP-completeness of Integer Fixpoint Computation

We now prove that the propagators $L_{i,k}$ introduced in the previous sub-section (Eq. 5), although very simple when considered independently, give rise to complex fixpoints. More precisely, we show the NP-completeness of the following decision problem:

**Decision Problem 1 (Bound(R)-Consistency for Linear Constraints)**
*INPUT: a CSP whose set of constraints $C$ are linear inequalities.*
*QUESTION: Let $F = \{L_{i,k} : i \in 1..n, k \in 1..m\}$ be the set of bound(R) consistency propagators associated with the CSP. Do the propagators in $F$ have a non-empty common fixpoint?*

#### 3.2.1 CHARACTERISING THE FIXPOINTS BY INEQUALITIES

Our first observation is that the bounds obtained when a fixpoint is reached are characterized by the bound consistency conditions of Eq. 4. In other words a fixpoint is reached iff the lower and upper bounds $x_i^-$ and $x_i^+$ satisfy the following inequalities, for each variable $i$ and constraint $k$:

$$\begin{cases} a_{1,k}x_1^{+s_{1,k}} + .. + a_{i-1,k}x_{i-1}^{+s_{i-1,k}} + \boxed{a_{i,k}x_i^{-s_{i,k}}} + a_{i+1,k}x_{i+1}^{+s_{i+1,k}} + .. + a_{n,k}x_n^{+s_{n,k}} \geq c_k & \\ & \forall k \in 1\ldots m, i \in 1\ldots n \quad (6)\\ l_i \leq x_i^- \leq x_i^+ \leq u_i & \forall i \in 1\ldots n \end{cases}$$

It is clear that Decision Problem 1 is answered positively iff there are *integer* values for the bounds $x_i^-$ and $x_i^+$, $i \in 1..n$, that satisfy the Linear Program 6. (If a fixpoint exists then the bounds given by this fixpoint satisfy the inequalities and are within the initial bounds $l_i, u_i$. Conversely if the inequalities are satisfied we have a fixpoint.)

A first consequence for Decision Problem 1 is that its membership in NP is straightforward since it is solvable by Integer Programming.

### 3.2.2 Linear Inequalities with Two-Variables-Per-inequality

The key to understanding why Decision Problem 1 is hard is to connect fixpoint computation to the special case of Integer (Linear) Programming where all constraints have *Two Variables Per Inequality* (*TVPI* in the LP terminology, see Bar-Yehuda & Rawitz, 2001):

**Definition 5** *A **TVPI** instance with $m$ constraints and $n$ variables is an Integer Linear Program of the following form:*

$$\begin{cases} a_k x_{i_k} + b_k x_{j_k} \geq c_k & \forall k \in 1\ldots m \\ l_i \leq x_i \leq u_i & \forall i \in 1\ldots n \end{cases}$$

*where $a, b, c$ are vectors of arbitrary (possibly negative) integers.*

The feasibility of TVPI constraints is NP-complete[3] but can be decided in pseudo-polynomial time. An early pseudo-polynomial time algorithm can be found in work by Aspvall and Shiloach (1980); this algorithm essentially reduces the problem to a 2-SAT instance of size $m \cdot d$, which is solvable in linear time (the overall algorithm therefore runs in pseudo-polynomial time, but also with a pseudo-polynomial space requirement). A particularly relevant algorithm for TVPI constraints is proposed in the work of Bar-Yehuda and Rawitz (2001). This algorithm has pseudo-polynomial time complexity with low, strongly polynomial space requirements. Interestingly, this algorithm essentially *uses bound propagation* (in fact, precisely bound(R) consistency), and embeds it in what amounts to a *backtrack-free* search with a "parallel" improvement that allows to amortize its overall runtime.

This seems to suggest a strong relation between propagation and TVPI constraints; in particular one could easily be mistaken to believe that propagation is a *decision procedure* for systems of TVPI constraints. We say that propagation provides a decision procedure for a class of constraints if propagation fails exactly when the constraints are unsatisfiable (in

---

3. Here we focus on feasibility only. The optimization problem, i.e., optimizing a linear function under TVPI constraints, is *strongly* NP-hard, i.e., NP-hard even for bounded domain sizes (in fact domains $\{0, 1\}$ are enough), because it trivially encodes Max-2SAT (Bar-Yehuda & Rawitz, 2001).

other words: the existence of a bound consistent state suffices to guarantee the existence of a solution). This is the usual condition that guarantees a backtrack-free search; but propagation rarely achieves this in the general case and it is in fact *not* a decision procedure for TVPI constraints:

**Example 2** *Consider the problem $x + y = 1, x = y$ with $x, y \in [0, 1]$. The problem is inconsistent yet it is bound(R) consistent (and also, in fact, bound(Z) consistent).*

To prove our main result we need to identify a restricted case of TVPI constraints for which fixpoint computation is indeed a decision procedure. This particular case is *monotone* TVPI constraints, in which the two variables in each inequality have coefficients with opposite signs, i.e., the problem is of the following form:

**Definition 6** *A **monotone TVPI** instance with $m$ constraints and $n$ variables is an Integer Linear Program of the following form:*

$$\begin{cases} a_k x_{i_k} - b_k x_{j_k} \geq c_k & \forall k \in 1 \ldots m \\ l_i \leq x_i \leq u_i & \forall i \in 1 \ldots n \end{cases}$$

*where $a_k \geq 0, b_k \geq 0, \forall k \in 1 \ldots m$.*

We can now prove our NP-hardness result from monotone TVPI constraints, using the following result:

**Theorem 1** *(Lagarias, 1985) The feasibility of Two-Variable-Per-Inequality* monotone *Integer Programming is NP-complete.*

### 3.2.3 NP-HARDNESS

We now prove that Decision Problem 1 is NP-hard. We already know that it is in NP, therefore we can state our main result for bound(R) consistency for linear constraints as:

**Proposition 1** *Decision problem 1 is NP-complete.*

**Proof.** *We show that fixpoint computation decides systems of monotone TVPI constraints. Consider a monotone TVPI instance $\mathcal{Q}$ of the form given by Def. 6. We want to show the equivalence: $\mathcal{Q}$ has an integer solution iff the set of bound(R) consistency propagators obtained for $\mathcal{Q}$ have a non-empty common fixpoint.*

- *"$\mathcal{Q}$ has an integer solution" means that there exist integer values $v_i$ for each variable $x_i$ satisfying $l_i \leq v_i \leq u_i$ and, $\forall k \in 1 \ldots m$:*

$$a_k v_{i_k} - b_k v_{j_k} \geq c_k \tag{7}$$

- *"There exists a common fixpoint" means that bounds $x_i^-$, $x_i^+$, can be found for all $i \in 1 \ldots n$, satisfying $l_i \leq x_i^- \leq x_i^+ \leq u_i$ and, $\forall k \in 1 \ldots m$:*

$$a_k x_{i_k}^- - b_k x_{j_k}^- \geq c_k \qquad a_k x_{i_k}^+ - b_k x_{j_k}^+ \geq c_k \tag{8}$$

*(These are simply the constraints of Eq. 6 for variable $x_{i_k}$ (left) and $x_{j_k}$ (right), rewritten by taking into account $a > 0, b > 0$.)*

*We prove the two directions of the iff:*

- *Consider an integer solution to $\mathcal{Q}$ in which each variable $x_i$ takes value $v_i$. It is easy to verify that the bounds $x_i^- = x_i^+ = v_i$ satisfy $l_i \leq x_i^- \leq x_i^+ \leq u_i$ and Eq. 8.*

- *Consider a bound consistent state described by the bounds $x_i^-$, $x_i^+$. It is easy to verify that the solution $v$ defined by $v_i = x_i^+, i \in 1 \ldots n$ satisfies $l_i \leq v_i \leq u_i$ and Eq. 7.*

*This means that we can reduce the problem of monotone TVPI feasibility to the existence of a fixpoint, and that Decision Problem 1 is therefore NP-hard.* $\quad\square$

Note that the NP-hardness result for Decision Problem 1 holds *even for (monotone) TVPI* constraints, while the pseudo-polynomial upper bound of Section 2.4 holds for general linear constraints of unbounded sizes; as said earlier the membership in NP is also valid for general linear constraints.

### 3.3 A Comment on Linear Equalities

From the beginning of this section we have focused on linear *inequalities* for reasons that should become clear in this sub-section. Readers may wonder whether considering *equalities* would make any difference. The short answer is *no*.

A first observation is that an inequality $\sum_{i \in 1 \ldots n} a_i x_i \geq c$ can directly be encoded into the equality $\sum_{i \in 1 \ldots n} a_i x_i - y = c$ where $y$ is a new variable ranging over $[0, u]$ for $u > \sum_i a_i x_i^{+s_i}$, so that there is a bijection between solutions of the two constraints. Therefore the problem of propagating inequalities reduces to the problem of propagating equalities, and the NP-completeness result still holds for problems whose linear constraints are all equalities (or any mix of equalities / inequalities).

A second observation is the following: because we focus on bound(R) consistency, the propagation obtained for an equality $\sum_{i \in 1 \ldots n} a_i x_i = c$ is the same as the one obtained using two constraints $\sum_{i \in 1 \ldots n} a_i x_i \leq c$ and $\sum_{i \in 1 \ldots n} a_i x_i \geq c$. For this reason it is convenient to assume that constraints are of homogeneous form, and to restrict ourselves to linear inequalities[4].

### 3.4 Tractable Classes of Linear Constraints

Intractable problems often become tractable when additional restrictions are imposed on the topology of the constraint graph, or on the constraints themselves. In this subsection we identify one significant class of linear constraints that can be propagated in strongly polynomial time, based on a restriction on the *coefficients* of the constraints.

Our initial observation is that one source of complexity of the propagators $L_{i,k}$ of Eq. 5 is that they use *rounding*: when we update a variable's bounds, we obtain from the other variables a real value that is rounded upwards for lower bounds and downwards for upper bounds. The effects of rounding were noticed by previous authors and used to optimize propagation (Harvey & Stuckey, 2003). Rounding effectively means that propagation stabilizes to *integral* solutions of Linear Programs. The Linear Programs in question have a very specific form, but the intractability is due to the integrality. Therefore in this sub-section (1) we observe that if we remove the rounding, the problem becomes tractable; (2) we use this observation to show that if the coefficients are unit (i.e., belong to $\{-1, 0, +1\}$), there is effectively no rounding, which means that the same tractability result holds.

---

4. Note also that in the case of inequalities the propagators for bound(Z) consistency are the same as for bound(R) consistency. Since we want propagators to be polynomial-time computable, the case we want to avoid however is *bound(Z) consistency for linear equalities*, where we cannot define polynomial-time computable propagators unless P=NP.

### 3.4.1 Linear Propagators without Rounding

We now consider operators similar to those of Eq. 5 but without rounding, in other words we now associate to the linear constraints the following operators:

$$\boxed{S_{i,k}:} \quad \begin{aligned} &\text{if } a_{i,k} > 0 \text{ then } \quad x_i^- \quad := \quad \max\left(x_i^-, \frac{q_{i,k}}{a_{i,k}}\right) \\ &\text{if } a_{i,k} < 0 \text{ then } \quad x_i^+ \quad := \quad \min\left(x_i^+, \frac{q_{i,k}}{a_{i,k}}\right) \end{aligned} \tag{9}$$

Even when the initial bounds are integers as assumed throughout this paper, these operators will in general reduce these bounds to real-values. Note that such propagators can effectively be used to deal with variables with a real-valued domain, indeed they are used both in the Constraint Programming community (Behamou & Granvilliers, 2006) and in the Operations Research community, where a different terminology is used (*Feasibility-Based Bounds Tightening*, see e.g. Belotti, Cafieri, Lee & Liberti, 2010).

The decision problem we focus on is now whether there exist real-valued bounds that are a fixpoint. We note that this problem is tractable; a similar result has been reported independently in the work of Belotti, Cafieri, Lee, and Liberti (2010).

**Decision Problem 2 (Fixpoint of Continuous Linear Propagators)**
*INPUT: a CSP whose set of constraints $C$ are linear inequalities.*
*QUESTION: Does the set of real-valued propagators $F = \{S_{i,k} : i \in 1..n, k \in 1..m\}$ associated to $C$ have a common fixpoint?*

**Observation 2** *Decision problem 2 can be decided by Linear Programming.*

It is easy to see that the fixpoints of operators $S_{i,k}$ are exactly the *real-valued* solutions to the system of linear constraints of Eq. 6. Note that we have been careful in the statement of Observation 2: whether Linear Programming is *strongly* polynomial is in fact a long-standing open question (Smale, 1998). The best "polynomial-time" LP algorithms are, encouragingly, of time complexity $\mathcal{O}(\pi(n, m, b))$ for some polynomial $\pi$, where $b$ is the number of bits of the number encoding—this looks strongly polynomial (Khachian, 1979). But there is a catch: the complexity is counted in number of operations, and operations on the rationals can in principle expand the size of the numbers (repeated multiplications can blow-up the representation exponentially). However, for practical purposes, typical LP implementations prevent the blow-up of number representation by limiting the precision to $b$ bits throughout the execution; solvability by Linear Programming is widely regarded as synonymous to strong tractability, and provably sub-exponential LP algorithms exist (Matousek, Sharir, & Welzl, 1996). In other words, Observation 2 should really be read as a carefully phrased way to say that Problem 2 is efficiently solvable in practice.

### 3.4.2 Linear Constraints with Unit Coefficients

A *unit* linear constraint is of the usual form $\sum_{i \in 1...n} a_{i,k} x_i \geq c_k$ but with the additional restriction that each coefficient $a_{i,k}$ is chosen in $\{-1, 0, +1\}$. Our introductory example of slow convergence (Eq. 1) was a (particularly simple) example of unit linear constraints, and the slow convergence could in this particular case be avoided. Note that we are considering linear unit constraints of *any number of variables*. A special case of unit constraints that have been widely studied is the class of unit-TVPI constraints (i.e., both unit *and* TVPI). This is perhaps the most important class of linear constraints whose integer feasibility can be solved in strongly polynomial time, see for instance work by Jaffar et al. (1994).

**Proposition 2** *When all constraints have unit coefficients, Decision Problem 1 can be decided by Linear Programming.*

**Proof.** *The LP is, of course, of the form given in Eq. 6. The observation is, in short, that* no rounding is needed *when the coefficients are unit.*

*More precisely, for any Cartesian product of intervals $A$, let $L(A) = \bigcap_{i,k} L_{i,k}(A)$ and $S(A) = \bigcap_{i,k} S_{i,k}(A)$. We show that when all coefficients are unit and when (as defined) the bounds of the initial Cartesian product $\mathcal{D}$ are integral, then we have $L^t(\mathcal{D}) = S^t(\mathcal{D})$, for all $t \geq 0$. We first note that the bounds of $L^t(\mathcal{D})$ are integral for all $t$ since the original state $\mathcal{D}$ has integral bounds and that each operator in $L$ applies rounding. The equality $L^t(\mathcal{D}) = S^t(\mathcal{D})$ is now proved by induction on $t$. For $t = 0$, $L^t(\mathcal{D}) = S^t(\mathcal{D}) = \mathcal{D}$. If the induction hypothesis holds at step $t$, then $S^{t+1}(\mathcal{D}) = S(S^t(\mathcal{D})) = S(L^t(\mathcal{D}))$. Now $S^{t+1}(\mathcal{D}) = L(L^t(\mathcal{D})) = L^{t+1}(\mathcal{D})$ because $L^t(\mathcal{D})$ has integral bounds, hence applying $S$ or $L$ on this Cartesian product gives the same result. (In Eq. 5 all $q_{i,k}$s are integral in this case and all $a_{i,k}$s are unit therefore the division $q_{i,k}/a_{i,k}$ gives an integer, which means that the propagators $L_{i,k}$ with rounding return the same result as the non-rounded propagators $S_{i,k}$ of Eq. 9.)*

*Now having $L^t(\mathcal{D}) = S^t(\mathcal{D})$, for all $t \geq 0$ it is easy to see that $\text{gfp}\{L_{i,k}\} = \text{gfp}\{S_{i,k}\}$. Because the domains are finite $L^t(\mathcal{D})$ stabilizes for a finite $t$. For this particular $t$, $L^t(\mathcal{D})$ is the greatest fixpoint of $L$ and the same greatest Cartesian product $S^t(\mathcal{D})$ is also the greatest fixpoint of $S$.* $\square$

Note that in general Linear Programming does not necessarily find integer solutions to the system of Eq. 6; what the result shows is that *LP will find a solution iff an integer one exists*. If we want to actually compute the largest consistent bounds $x_i^-$ and $x_i^+$ of a certain variable $x_i$, we can simply minimize $x_i^-$ or maximize $x_i^+$ under the constraints of Eq. 6. The previous proof shows that these extremal values are integral.

### 3.4.3 Are There Other Tractable Cases?

It is interesting to consider whether other properties make the propagation solvable in strongly polynomial time. With respect to restrictions on the constraint graph, there are nevertheless reasons to be pessimistic: we note that the feasibility of monotone TVPI Integer Programming remains NP-complete under strict restrictions on the constraint graph, as shown in the work of Hochbaum and Naor (1994). This suggests that such restrictions are unlikely to lead to interesting tractable classes of fixpoint computation.

Regarding the restrictions on coefficients, we note that in general the NP-completeness of (monotone) TVPI constraints assumes that the coefficients $a_k, b_k, c_k$ are arbitrary. The *Unit* restriction imposes, on the contrary, the strongest restriction on coefficients: that their absolute value be $\leq 1$. If we impose a more general bound $\beta$ on these absolute values then one may wonder whether the problem exhibits some form of *fixed-parameter tractability*. We leave this question open for future work.

## 4. Generalizations and Non-Linear Constraints

By Proposition 1, fixpoint computation for numerical constraints as basic and common as linear constraints is intractable. Several cases of non-linear constraints are nevertheless of interest. First, we show that if the simplest possible type of polynomials (a single squaring operation) is added to linear constraints, then our general hardness result can be strengthened. Second, it is interesting to note that if we enrich *unit* linear constraints with simple min or max constraints, then fixpoint computation is equivalent to a puzzling open problem discussed recently in the theorem-proving literature. Last, we briefly comment on connections between our results and the tractability of max-*closed* constraints.

### 4.1 Quadratic Constraints

For the purposes of this section it is sufficient to enrich our linear constraint language (constraints of the form given by Eq. 2) with squaring constraints of the form:

$$x_i = x_j^2$$

It is also sufficient to restrict ourselves to non-negative values for variables $x_i$ and $x_j$, i.e., $0 \leq l_i \leq u_i$ and $0 \leq l_j \leq u_j$. In this setting the bound(R) consistency propagators are defined by the following instructions:

$$x_i^- := \max(x_i^-, \left(x_j^-\right)^2) \qquad x_i^+ := \min(x_i^+, \left(x_j^+\right)^2)$$

$$x_j^- := \max(x_j^-, \left\lceil \sqrt{x_i^-} \right\rceil) \qquad x_j^+ := \min(x_j^+, \left\lfloor \sqrt{x_i^+} \right\rfloor)$$

In other words the fixpoints are integer solutions to the following bound consistency inequalities:

$$x_i^- \geq \left(x_j^-\right)^2; \quad x_i^+ \leq \left(x_j^+\right)^2; \quad x_j^- \geq \sqrt{x_i^-}; \quad x_j^+ \leq \sqrt{x_i^+} \tag{10}$$

When these simple quadratic constraints are added to the language of linear constraints, our NP-completeness result can be strengthened: the problem is NP-complete even when considering a *bounded*(!) number of variables and constraints; in fact one TVPI constraint and one squaring constraint. This is due to the fact that fixpoint computation converges to a state that encodes a complex number-theoretic problem.

**Proposition 3** *Given a CSP with 3 variables and 2 constraints $a_1 x_1 + a_2 x_2 = c, x_1 = x_3^2$, determining whether their associated bound(R) consistency propagators have a fixpoint is NP-complete.*

**Proof.** *Membership in NP is straightforward. We show the hardness result for the special case where $a_i \geq 0, i \in \{1, 2, 3\}$ and focus, as said, on positive intervals. We first note that the bound consistency inequalities of (Eq. 10) for the squaring constraint $x_1 = x_3^2$ are satisfied iff $x_1^- = (x_3^-)^2$ and $x_1^+ = (x_3^+)^2$ since we focus on integer bounds. (This property of the squaring propagator is noticed in a slightly different form in Schulte & Stuckey, 2005). From a propagation viewpoint the equality $a_1 x_1 + a_2 x_2 = c$ is seen as two inequalities $a_1 x_1 + a_2 x_2 \geq c$ and $-a_1 x_1 - a_2 x_2 \geq c$ whose bound consistent inequalities (Eq. 4) are effectively satisfied iff $a_1 x_1^- + a_2 x_2^+ = c$ and $a_1 x_1^+ + a_2 x_2^- = c$.*

*We rely on a theorem (Manders & Adleman, 1978) which shows that deciding whether an equation of the form $a_1 x_3^2 + a_2 x_2 = c$ has integer solutions, where $a_1$, $a_2$ and $c$ are non-negative integers, is NP-complete. We reduce this problem to the existence of bound consistent bounds for the conjunction $a_1 x_1 + a_2 x_2 = c, x_1 = x_3^2$ with initial bounds $l_1 = l_2 = l_3 = 0$ and $u_1 = u_2 = u_3 = c$. We just need to show that fixpoint computation is complete for this system—a bound consistent state is found iff the original equation has a solution:*

- *If the original equation has a solution, i.e., a pair of non-negative integer values $\langle v_2, v_3 \rangle$ satisfying $a_1 v_3^2 + a_2 v_2 = c$, then we define $x_1^- = x_1^+ = v_3^2$, $x_2^- = x_2^+ = v_2$, and $x_3^- = x_3^+ = v_3$. These bounds are such that $0 \leq x_i^- \leq x_i^+ \leq c$ and satisfy the bound consistency conditions of Eq. 10 and Eq. 4.*

- *If the conjunction has a bound consistent state, i.e., bounds $x_i^-, x_i^+$ such that Eq. 10 and Eq. 4 are satisfied, then the solution $v$ defined by $v_2 = x_2^+$ and $v_3 = x_3^-$ satisfies the original equation $a_1 v_3^2 + a_2 v_2 = c$.*

□

## 4.2 Connections to the Max-Atom Problem

Another common type of primitive non-linear constraints is of the form:

$$x_h = \max(x_i, x_j)$$

The bound(R) consistency propagators for this constraint are the following (Schulte & Stuckey, 2005):

$$x_h^- := \max(x_h^-, x_i^-, x_j^-) \qquad x_i^+ := \min(x_i^+, x_h^+)$$

$$x_h^+ := \min(x_h^+, \max(x_i^+, x_j^+)) \quad x_j^+ := \min(x_j^+, x_h^+)$$

(In fact to strictly reach bound(R) consistency one would need to additionally check whether the bounds of $x_h$ have an empty intersection with the bounds of one of the max-ed variables, say $x_i$, in which case we can essentially impose the constraint $x_j = x_h$; for the purposes of this section the simpler formulation above is equivalent.) In other words the fixpoints are characterized by the following inequalities:

$$x_h^+ \leq \max(x_i^+, x_j^+) \qquad x_i^+ \leq x_h^+ \qquad x_j^+ \leq x_h^+ \qquad x_h^- \geq x_i^- \qquad x_h^- \geq x_j^-$$

The fixpoint computation of max constraints mixed with *unit* linear constraints is interesting because its complexity is an open problem. Note that there is no rounding or use of coefficients in the definition of the bound consistency inequalities, therefore the complexity arising from rounding in all our NP-complete variants of propagation does not arise here. The open problem we connect to is called *Max-Atom* in the work of Bezem, Nieuwenhuis, and Rodríguez-Carbonell (2008); see this reference for prior problems of interest that are shown equivalent to Max-Atom. A max-atom constraint is of the form: $\max(x_i, x_j) + c \geq x_h$. The work reported by Bezem et al. (2008) shows a number of results on the feasibility of conjunctions of max-atom constraints: (1) There is no significant complexity difference between integer and real feasibility; (2) The problem can be decided in pseudo-polynomial time using what amounts to a fixpoint computation algorithm; (3) The problem has short proofs of unsatisfiability and is therefore in NP∩coNP; which means that it is of a very different nature from our other NP-complete variants. In fact, a recent result (Atserias & Maneva, 2010) shows that the complexity of Max-Atom is equivalent to well-known open problems called *mean-payoff* games, which have in turn connections to some important open questions in model-checking: parity games, a class of games reducible to mean-payoff games, are equivalent to the model-checking problem of $\mu$-calculus (Emerson, Jutla, & Sistla, 1993; Jurdzinski, 1998).

Here we draw a simple connection that follows from the observation that the bound consistency inequalities for the upper bounds include the constraint $x_h^+ \leq \max(x_i^+, x_j^+)$ which encode max-atom constraints almost directly.

**Proposition 4** *Bound(R) consistency for a combination of unit linear and* max *constraints can be solved in polynomial time only if Max-Atom can be also be solved in polynomial time.*

**Proof.** *To reduce a Max-Atom instance with variables* $x_i, i \in 1 \ldots n$ *and* $m$ *constraints to a fixpoint computation problem we simply introduce one fresh variable* $y_k$, *for each* $k \in 1 \ldots m$. *Let the* $k$th *constraint be of the form* $\max(x_{i_k}, x_{j_k}) + c_k \geq x_{h_k}$, *it rewrites to the conjunction* $\max(x_{i_k}, x_{j_k}) = y_k, y_k + c_k \geq x_{h_k}$. *The lower bounds of all variables are fixed to 0 and the upper bounds need only be set to* $\sum_{k \in 1 \ldots m} c_k$ *by the small model property (Lemma 2) of the paper by Bezem et al. (2008). The bound consistency equations for the upper bounds directly encode the problem.* □

## 4.3 Max-Closed Constraints

We last note a connection between our results and the class of max-closed constraints introduced by Jeavons and Cooper (1995) (more on this in, e.g., Petke & Jeavons, 2009). A constraint $R(x_1, \ldots, x_n)$ is max-closed if whenever we have two solutions $\langle v_1 \ldots v_n \rangle$ and $\langle w_1 \ldots w_n \rangle$, their maximum defined as $\langle \max(v_1, w_1), \ldots, \max(v_n, w_n) \rangle$ is also a solution. Results by Jeavons and Cooper (1995) show that max-closed constraints are tractable: if a system of constraints is max-closed, then its feasibility can be determined in polynomial time. However note that this result essentially assumes an *explicit* (or table) representation of the constraint, i.e., it is assumed that each constraint is defined by explicitly listing the tuples that are solutions to it. In contrast some important types of constraints such as the numerical constraints considered in this paper are *implicitly defined*: we do not know the list of solutions to $R(x_1, \ldots, x_n)$ but can only verify efficiently whether a particular tuple is accepted by the constraint.

Implicitly-defined max-closed constraints played an important role in this paper: both the monotone TVPI constraints, considered in Section 3.2 and the Max-atom constraints considered in Section 4.2, are max-closed, as shown respectively by Hochbaum and Naor (1994), and Bezem et al. (2008). In sharp contrast to the case of explicitly-defined constraints, the resolution of implicitly-defined max-closed constraints is therefore only pseudo-polynomial and it is in fact intractable, as shown by the special case of monotone TVPI constraints:

**Observation 3** *The feasibility of "implicitly-defined" max-closed constraints is NP-complete.*

As shown in Section 3.2 with the particular example of monotone TVPI constraints, *even the fixpoint computation* of implicitly defined max-closed constraints is, in fact, NP-complete in general.

## 5. Conclusion

Reasoning about intervals was introduced in the AI literature by the works of Cleary (1987), and Davis (1987)[5]. A substantial body of AI work has ensued (see, e.g. Hyvönen, 1992); bound computation is now used by most finite-domain CP solvers (Schulte & Carlsson, 2006).

In this paper we have theoretically investigated the complexity of computing the common fixpoint of a set of bound consistency propagators. We have shown that even when the propagators are themselves very simple, the fixpoint computation used in these algorithms can be complex, it is indeed NP-complete even for a very restricted constraint class – linear monotone inequalities with two variables per inequality. We also considered some special classes of constraints, like quadratic constraints and max constraints. Finally, we identified a class of constraints, namely, linear inequalities with unit coefficients, that allows a tractable fixpoint computation algorithm.

Bound propagation is a successful and widely used technique in Constraint Programing. There is a large literature on propagating single constraints (Van Hoeve & Katriel, 2006; Bessière, 2006; Rossi, van Beek, & Walsh, 2006) and it is perhaps a surprise that no prior study exists on the complexity of the fixpoint computation. The NP-completeness of fixpoint computation for simple types of constraints is a fundamental and somewhat surprising result, and one that sheds light on slow convergence phenomena.

This result also "puts bound propagation on the map" of AI computational problems: together with knapsack constraints and some forms of learning in neural nets (Schaeffer & Yannakakis, 1991), it is one of the few important AI problems we are aware of that have a pseudo-polynomial complexity and yet are intractable.

### Acknowledgments

Preliminary results of Bordeaux, Hamadi, and Vardi (2007) showed the NP-completeness of propagation in the case where quadratic constraints are considered (Prop. 3). This paper is a thoroughly revised version whose central result for linear constraints is new and more general. Part of this work was done while G. Katsirelos, N. Narodytska and M. Vardi were visiting Microsoft Research, Cambridge. Part of this work was done while G. Katsirelos was employed by NICTA, Australia. NICTA is funded by the Australian Governments Department of Broadband, Communications, and the Digital Economy and the Australian Research Council. This work was partially supported by the ANR UNLOC project: ANR 08-BLAN-0289-01. Discussions with Youssef Hamadi and Claude-Guy Quimper are gratefully acknowledged. Thanks also to the anonymous reviewers whose feedback helped improve the paper.

---

5. As often, a good case can be made that similar ideas are already present in earlier work, in particular the work of Laurière (1978). Interval computations are of course also used in other areas and the fixpoint computation methods we consider relate to the broader theme of *interval arithmetic* pioneered by Moore (1966).